**Representation in large language models**

Cameron Yetman | cameron.yetman@mail.utoronto.ca
University of Toronto

**Abstract**
The extraordinary success of recent Large Language Models (LLMs) on a diverse array of tasks has led to an explosion of scientific and philosophical theorizing aimed at explaining how they do what they do. Unfortunately, disagreement over fundamental theoretical issues has led to stalemate, with entrenched camps of LLM optimists and pessimists often committed to very different views of how these systems work. Overcoming stalemate requires agreement on fundamental questions, and the goal of this paper is to address one such question, namely: is LLM behavior driven partly by representation-based information processing of the sort implicated in biological cognition, or is it driven entirely by processes of memorization and stochastic table look-up? This is a question about what kind of algorithm LLMs implement, and the answer carries serious implications for higher level questions about whether these systems have beliefs, intentions, concepts, knowledge, and understanding. I argue that LLM behavior is partially driven by representation-based information processing, and then I describe and defend a series of practical techniques for investigating these representations and developing explanations on their basis. The resulting account provides a groundwork for future theorizing about language models and their successors.

## 1. Introduction

Artificial intelligence systems have played at least two major roles in the development of cognitive science. First, they have been used as models of cognition – toy systems which help us conceptualize the kinds of processes which underlie our ability to plan, reason, navigate, and understand the world (Newell 1980; Rumelhart & McClelland 1986; Fodor & Pylyshyn 1988; Stinson 2020; Griffiths et al. 2024). Second, they have been treated as candidate cognitive systems themselves – as not only modeling cognition, but instantiating it (Turing 1950; McCarthy 1979; Aleksander 2001; Chella & Manzotti 2007; Goldstein & Levinstein 2024). These roles are connected, since a perfect model of cognition is simply one which replicates cognitive processes, but a system which replicates cognitive processes is precisely a system which *has* cognitive processes, at least given some kind of functionalism.[1]

Large language models (LLMs) play both of these roles in the recent history of cognitive science. Some papers treat them as potential models of cognition (Jin et al. 2022; Binz & Schulz 2023; Blank 2023; Niu et al. 2024), while others debate whether they themselves can reason, understand, or communicate meaningfully (Bender & Koller 2020; Bubeck et al. 2023; Dhingra et al. 2023; Mitchell & Krakauer 2023; Goldstein & Levinstein 2024; Stoljar & Zhang 2024). Unfortunately, interlocutors often come to these debates with very different understandings of what cognition involves (and so what it takes to model or instantiate it) and of what kinds of processes actually underlie the behavior of LLMs. At one extreme are those for whom LLMs' success on apparently cognitive tasks entails the possession of cognitive capacities. On the other are those for whom no amount of success has this consequence, since the way in which LLMs achieve it is incompatible with their being genuine cognizers (Marcus & Davis 2020; Bender et al. 2021; Marcus 2022; Titus 2024). In particular, some have argued or assumed that LLMs are little more than stochastic look-up tables. These disagreements are quickly becoming entrenched into camps of LLM optimists and pessimists, and accordingly we are at risk of disciplinary stalemate.

[1] That is, the view that mental states are defined by their functional role within a cognitive system: Putnam (1960, 1967); Armstrong (1968); Block & Fodor (1972); Lewis (1972); Churchland (2005); Cao (2022a); Levin (2023). Chalmers (1996) frames the twofold role of computational systems in cognitive science nicely: "it is hoped that computation will provide a powerful formalism for the *replication* and *explanation* of the mind" (309).

Overcoming stalemate requires agreement on the fundamentals. The goal of this paper is to reconsider one fundamental issue in the background of current debates, namely the question of whether LLM behavior is driven by representation-based information processing, or whether it is driven by non-representational processes of memorization and stochastic table look-up. Representation-based information processing is typically considered the foundation of human and animal cognition (Fodor 1975; Sterelny 1991; Eckardt 2012; Shea 2018); the algorithms posited to explain our cognitive capacities are formulated in representational language. Although there is persistent disagreement over their nature and extent (Clark & Toribio 1994; Van Gelder 1995; Chemero 2000; Clark 2015; Constant et al. 2021), *that* representations play a central role in (especially higher) cognitive processes such as language use, reasoning, and abstract problem solving is reasonably uncontroversial, and so should serve as common ground for both sides of the LLM debate. Language use, reasoning, and problem solving are all tasks at which LLMs have shown surprising proficiency (Gubelmann et al. 2024; Liu et al. 2023; Naik et al. 2023).[2] If they perform these tasks via representational means, they are more likely to serve as adequate cognitive models, and more likely to be candidate cognizers themselves. If, on the other hand, they produce outputs through non-representational processes of memorization and stochastic table look-up, they are less plausible candidates for filling either of these roles.

A number of recent and upcoming papers assume a representationalist framework for explaining LLM behavior. These explanations operate roughly at Marr's "algorithmic level": they describe how the rulebound interactions of internal information-structures (i.e., representations) lead from input to output (Marr 1982; Millière & Buckner 2026).[3] Different explanations posit distinct rules and interactions, but representations are part of the overall explanatory story. For instance, circuit-based analyses describe how the computational pathways connecting represented features in LLMs give rise to behavior (Elhage et al. 2021; Dunefsky et al. 2024), while causal abstraction analyses aim to align the posited internal representations of a target model with the nodes of a lower-level causal model of behavior (Geiger et al. 2021; Budding & Zednik 2024).

---

[2] For reviews of LLM behavior, see Zhuang et al. (2023); Chang & Bergen (2024).

[3] Though see Adolfi et al. (2024) for some in-principle limits to such explanations.

In defense of this framework, I argue that LLM behavior *is* driven by representation-based information processing, at least in some cases. My argument has the following steps. First, I propose a characterization of "representation" which captures their unique explanatory role in cognitive science (section 2). Next, I outline a popular non-representational strategy for explaining LLM behavior, and argue that it fails in some key cases, leaving representation-based explanations as the only viable alternative (section 3). Finally, I describe and defend the viability of a series of practical techniques for investigating LLM representations (section 4). Thus, the paper aims both to settle the question of *whether* LLMs behave in virtue of representational processes, and to describe *how* we can go about building explanations of their behavior on that basis.

The viability of representation-based explanation in the LLM context has recently received some attention from philosophers,[4] with the general conclusion being that representations are invaluable tools in our explanatory toolbelt. In this respect, my argument contributes to an emerging philosophical consensus. Its novelty consists in its greater precision concerning the conditions which must be satisfied for a state to count as a representation, its way of framing and rejecting the alternative explanatory strategy, and its explicit and thorough philosophical defense of the experimental techniques being used to investigate the representations of LLMs and other DNN-based AI systems.[5]

Note that I remain neutral on the questions of *what* LLMs represent (eg. merely features of text or features of the wider world), and of whether their representations constitute thoughts, concepts, beliefs, knowledge, or facilitate understanding. These higher-level questions are likely to be illuminated by my lower-level discussion, but are beyond the scope of this paper.[6]

## 2. Representation

In order to survive and flourish, organisms must solve the difficult task of appropriately modulating their behavior in response to environmental conditions. Sometimes this can be done quite directly, as in the case of reflexes, where there is an automatic stimulus-response pattern either learned from experience or built innately into the organism. Other times, organisms can take advantage of dynamic perception-action feedback loops in order to actively and continuously control the values of relevant internal or external variables (Van Gelder 1995; Chemero 2000). Yet other times, the task requires gathering information about features of the environment, forming internal proxies of those features, and processing those proxies over time in order to figure out what to do. These proxies are called "representations", and they plausibly undergird most forms of higher cognitive processing. Representations need not be conscious or introspectively accessible to the system which employs them; the paradigm of representation in this sense is not the conscious perceptual representation or person-level attitude, but the subconscious and subpersonal cognitive map, feature detector, or parse tree.

---

[4] See especially Harding (2023), Tamir & Shech (2023), Grzankowski (2024), and Williams et al. (2025)

[5] See section 2.5. for further discussion of how my view compares with that of Harding (2023).

[6] For discussion of each of these issues, see Stoljar & Zhang (2024) on LLM thinking, Mitchell & Krakauer (2023) on concepts and understanding, Herrmann & Levinstein (2025) on beliefs, and Yildirim & Paul (2024) on knowledge. For broader discussions of the possibility of LLM mentality, see Grzankowski et al. (2025) and Goldstein & Levinstein (2024). Naturally, all of these only scratch the surface of deep and ongoing debates.

The claim that LLMs or any other neural network (NN)-based system have representations as well is likely to strike some machine learning (ML) practitioners as trivial, since in ML the activity of NNs is often *defined* in terms of the formation of representations – as a kind of "representation learning" (Bengio et al. 2014). However, the term "representation" is used so liberally in ML practice that it no longer picks out an especially useful theoretical kind. As Ramsey (2017) has noted, essentially *any* causal intermediary between an NN's input and output is considered a representation, regardless of its properties: "[t]he representational theory of mind is slowly becoming the 'causally relevant to the processing theory of mind'—an utterly vacuous outlook" (2017: 4206). Accordingly, the aim of this section is to defend a characterization of representation which is substantial enough to capture its special explanatory role in cognitive science. The characterization consists of four jointly sufficient conditions, though I remain neutral about whether they are necessary.[7,8]

**REPRESENTATION**

System S's internal state R represents feature *z* if:

| | |
|---|---|
| INFORMATION | R carries information about *z* |
| EXPLOITABILITY | The information R carries about *z* is exploitable by S |
| BEHAVIOR | S's exploiting the information R carries about *z* enables S's robust *z*-related behavior |
| ROLE | R plays a mechanistic role in S's robust *z*-related behavior[9] |

"System" is intentionally vague, and is intended to denote any organized physical entity (egs. a human, a computer, a fish, a neural network, etc.). Different systems will have different candidate Rs; paradigmatically, they are neural states and collections thereof, or states of electrical potential on a chip, but may take other forms as well. Feature *z* can, for the purpose of this paper, be anything, including an aspect of the environment or an aspect of S's input space.[10]

### *2.1. INFORMATION*

INFORMATION captures the idea that representations are proxies for the feature(s) represented. Intuitively, in order to think about something that isn't immediately present to you, you must have cognitive access to something else that stands for, or represents, that thing. A stand-in need not be a perfect or accurate model of the feature: I can think about Toronto without knowing everything about it. Rather, it need only carry information about the feature. The relevant sense of information is *mutual information*, which, colloquially, is a measure of the degree to which knowledge of the value of one variable reduces uncertainty about the value of another (Cover & Thomas 2006). For example, if X is a binary variable corresponding to the presence or absence of smoke, and Y is a binary variable corresponding to the presence or absence of fire, knowing the value of X (knowing whether there is smoke) reduces uncertainty about the value of Y (about whether there is fire). X and Y carry mutual information. Accordingly, one could use X as a proxy, or stand-in, for Y, for instance when deciding whether or not to deploy firefighters.

[7] Indeed, I am skeptical that a conceptual analysis of representation is possible.

[8] See Harding (2023) for a similar proposal, and Grzankowski (2024) for related discussion.

[9] Note that this is a characterization of thetic (mind-fits-world) representations, rather than telic (world-fits-mind) representations. Most discussions of thetic and telic representations are concerned with propositional attitudes like beliefs and desires, which I explicitly am not. However a similar distinction might plausibly apply even in the case of subpersonal, non-propositional representations (Shea 2018: ch. 7). If so, I am only concerned with the former, thetic kind of representation, insofar as these provide the system which has them cognitive access to the environment. See Humberstone (1992) for the thetic-telic distinction.

[10] Williams et al. (2025) has recently pointed out that the term "feature" is often used ambiguously in ML practice, sometimes to refer to the representational vehicle, other times to the representational content. In this paper, I use it in the content-sense,

This colloquial characterization is intuitive but unsatisfactory, since "knowledge" and "uncertainty" are mental state terms, and my account is meant to apply at a sub-mental (computational) level. More technically, then, we can define the mutual information *I* of two variables X and Y as the difference between the entropy H of X and the conditional entropy of X given Y:

(1) $I(X,Y) = H(X) - H(X|Y)$[11]

Entropy is a measure of the average surprise associated with sampling from a variable (i.e., with observing which value it takes). A variable with high entropy has high surprise, and low entropy low surprise. The conditional entropy of X and Y is a measure of how surprising X is *given* the value of Y. If X and Y are completely independent variables, such that observations of Y have no effect on observations of X, then $H(X|Y) = H(X)$, and so $I(X, Y) = 0$. If X and Y are not independent, $H(X|Y) < H(X)$, and so $I(X, Y) > 0$. INFORMATION is satisfied when $I(R, z) > 0$. That said, $I(R, z)$ is likely to be much higher in most actual cases, since resource-limited systems are unlikely to devote significant effort towards exploiting a proxy which conveys only a tiny amount of information about the feature of interest.[12]

Finally, there are different ways variables can become entwined such that they bear mutual information. Two ways which are especially important in cognitive science, as discussed in-depth by Shea (2018), are *causally generated correlations* between the variables (as in the smoke-fire case) and *structural correspondences* between them (as in the case of a map). I provide examples of both kinds throughout the paper.

### *2.2. EXPLOITABILITY*

EXPLOITABILITY says that the information R carries about *z* is exploitable by S. Intuitively, this means that S must be able to access or use the information. But "information" is not a substance, it cannot literally be used: it plays no causal role. Rather, S can exploit the information R carries about *z* when S's *z*-related behavior can be conditioned on the tokening of R in a content-relevant way.[13] Analogously, I can exploit the information my map of Toronto carries about the city's street layout when I can condition my navigational behavior on the map – i.e., when I can use it to navigate. Were I only able to use the map to, say, soak up spilled milk, I would not be able to use it in a content-relevant way, and so EXPLOITABILITY would not be satisfied. The reader may analyze the phrases "can exploit" and "be able to use it" according to their favourite theory of ability modals. But practically speaking, learning whether EXPLOITABILITY is satisfied in a particular case is likely to depend on the performance of tests and interventions which involve altering R to see if S's *z*-related behavior changes in appropriate ways given R's putative content. See section 4 for a detailed discussion of such tests in LLMs.

---

[11] See Cover & Thomas (2006). Note that (1) is equivalent to $I(X, Y) = H(Y) - H(Y|X)$. Thus, if X carries information about Y, Y also carries information about X. This symmetry would pose a problem if representation had *only* to do with carrying information, but the other conditions ensure this issue does not arise.

[12] See Harding (2023) for a detailed discussion of how to measure the information borne by the activations of a neural network.

### *2.3. BEHAVIOR*

BEHAVIOR requires that S's exploiting the information R carries about *z* enables S's robust *z*-related behavior. Two terms are worth clarifying here. First, "enables" is an intentionally weak requirement; my map of Toronto enables me to navigate, but so does wearing my glasses. However, any stronger requirement (eg. "guarantees", "ensures", etc.) would be too strong, since the way a representation influences behavior never depends merely on the representation, but also on the algorithm(s) and context(s) in which it is embedded: "even for a given fixed representation, there are often several possible algorithms for carrying out the same process. . . . [O]ne algorithm may be much more efficient than another, or another may be slightly less efficient but more robust" (Marr, 1982, 23-24). In other words, representations characteristically enable robust behavior *given* that they are embedded in an appropriate sort of algorithm, where appropriateness depends on the properties of the representing system and the demands of the task.

Second, a behavior is "robust" if it is insensitive to minor perturbations in surrounding conditions.[14] Robustness is a modal notion – a system must be insensitive to perturbations in different *possible* scenarios, even if they never actually encounter those scenarios. Plausibly, enabling modally robust behavior is one of the central functions of representations, and representation-based explanations are typically invoked precisely in order to explain such behavior. This is the key virtue of representation-based explanations over explanations in terms of more basic or lower-level goings-on. To borrow one of Shea's (2018) examples, we can provide a unified and perspicuous explanation of how a squirrel robustly reaches a bird feeder by attributing goal-representations and spatial-representations to the squirrel. If instead we try to explain its behavior in terms of basic physical or neurobiological activities, our explanation of how the squirrel reaches the feeder in one set of conditions will be very different from the explanation of how it reaches the feeder in another set, leaving us with a disjunctive rather than a unified account of its behavior. Of course, this claim has limits: a maximally unified account of all phenomena is "what happens, happens", but a good scientific explanation must also be an informative one.

---

[13] Cao (2022b) makes a similar point: "if you change the representations . . . upstream, you should also expect to change the behavior . . . downstream, and, crucially, in *intelligible ways relating to the putative content* of the representation" (151, emphasis mine).

[14] See Behrens et al. (2018) for a discussion of how map-like representations in particular enable robust behavior by encoding the structural relations between entities.

Some authors tie representations directly to behavioral *success*, rather than mere robustness: "correct and incorrect representation explains successful and unsuccessful behavior" (Shea 2018: 24).[15] This requirement is too strong, since the same representation embedded in different algorithms can enable either robust success *or* robust failure. Returning to our example, if I use my map of Toronto in the standard way – aligning the compass rose with my actual compass, turning right when the map tells me to turn right, etc. – it will directly enable my success at navigating. However, if I use it in a non-standard way – aligning the compass rose opposite to my actual compass, turning right only half of the time the map tells me to, etc. – it will directly enable my failure at navigating. The key point is that I will *robustly* succeed at navigating in the first case, and *robustly* fail in the second case. Tying representations to success would force us to say that I have a map in the first, but not the second case, which seems both ad hoc and obviously false.[16]

### *2.4. ROLE*

ROLE fleshes out the idea that S's *z*-related behavior is "conditioned on" the tokening of R. What it is for S's behavior to be conditioned on R is for R to play a mechanistic role in that behavior, where R plays a mechanistic role just in case it plays a causal role in the mechanisms driving the behavior (Craver 2007). This point helps combat worries about panrepresentationalism: since almost everything is correlated with everything else in some way, and hence carries information about those things, a characterization of representation should provide a principled way to carve out those which play the representation-role from the wider space of information-bearing entities and states, lest its theoretical value be neutered. For example, my map not only carries information about the layout of Toronto streets, but about the type of ink available at the factory where it was printed, about the formatting conventions of modern maps, etc. However, ROLE is only satisfied when the map plays a mechanistic role in my robust feature-related behavior; in the navigation example, I exhibit no behavior related to the proposition that such and such ink was available at the factory, or that such and such conventions are standard in modern maps, and so the map does not represent those features.

---

[15] See also Cao (2022b) and Herrmann & Levinstein (2025). Note that the latter are interested specifically in LLM *belief* representations, conditions on the attribution of which are both more and less specific than my own. Their conditions are more specific because they require that the feature about which internal state R conveys information is *truth* (whereas my account is not limited to any particular feature), and because they incorporate conditions of ACCURACY, COHERENCE, and UNIFORMITY which may plausibly constrain an account of belief, but not of representation in general. By contrast, their USE condition – which requires that "the LLM tend to use the identified representations in a role appropriate for belief to determine what probability distribution to output, or what text to produce" (18) – is a less specific version of my BEHAVIOR and ROLE conditions, which identify a particular sort of role, namely a mechanistic one, and a particular sort of behavior that a state playing this role should support, namely robust feature-related behavior. Hence, their account is best seen as specifying conditions on a particular *subtype* of representational state, namely beliefs, while their USE condition would benefit from precisification along the lines of BEHAVIOR and ROLE.

[16] See section 2.5. for further discussion.

This isn't to say it never could do so, but the conditions under which it could would need to be significantly different. Hence, panrepresentationalism is avoided.

The above characterization of representation is far more substantial than that implicit in machine learning practice, and it captures the most important features of representations as discussed in cognitive science – they are proxies for features of some environment or input space in virtue of the exploitable information they carry, and they play a causal role in enabling robust feature-related behavior. If LLMs turn out to have representations of this sort, that is no trivial matter; it would open up new avenues for understanding and explaining their behavior, and justify the application of existing theoretical and experimental tools from representation-based approaches to cognitive science. Before considering an alternative strategy for explaining LLM behavior according to which they are non-representational systems equivalent to look-up tables, it will be worth explaining how my account of representation compares to a prominent recent account due to Harding (2023).[17]

### *2.5. Comparison with Harding (2023)*

In "Operationalizing representation in natural language processing", Harding (2023) proposes three criteria she takes to characterize a useful notion of representation for NLP practice. Her criteria are as follows, where $Z$ is a property, $s$ is an input, and $D$ is a task:

> **The Information Criterion:** The pattern of activations $h(D)$ bears information about $Z$.
>
> **The Use Criterion:** The information the pattern of activations $h(D)$ bears about $Z$ is used by system $S$ to perform task $D$.
>
> **The Misrepresentation Criterion:** It should be possible (at least in principle) for the activation vector $h(s)$ to misrepresent $Z$ with respect to an input $s \in D$.

Harding devotes some space to elaborating on and defending these criteria, but spends most of the paper developing an operationalization of them in terms of probing classifiers and other techniques from the literature on *mechanistic interpretability*. I pursue a similar tack in section 4, although with less formal detail and greater attention to certain conceptual issues related to heuristics and the explanatory role of representations.

While our works are generally best seen as complementary (indeed, our INFORMATION conditions are nearly identical, and the use condition is closely related to BEHAVIOR and ROLE), there are at least three notable differences between them. The first is that Harding is interested in representations of "properties" (such as part-of-speech or gender properties), whereas my account applies more generally to "features", a non-specific class encompassing properties, objects, concepts, propositions, and possibly much else besides. This is in keeping with machine learning practice – which is rife with talk of features – and with a commitment to neutrality regarding the possible contents of LLM representations.

---

[17] Thanks to an anonymous reviewer for encouraging me to clarify this.

More significantly, Harding's criteria are task specific, insofar as the representational status of a set of activations depends on its informational and causal role relative to a particular task D. In effect, the satisfaction of her criteria deliver the verdict that "system S represents$_D$ feature *z*", where the subscript indicates the task to which the representation-attribution is relativized. This aligns with Harding's "operationalization first" (6) approach, since any particular set of activations which we are investigating will be tokened in a specific task setting, the outcome of which we are trying to explain.

However, incorporating task-specific considerations within the *criteria* for representation begets a significant reduction in the generality of our representation-attributions, which carries an explanatory cost. The fact that S represents *z* should be relevant to explaining S's behavior across a *range* of tasks to which *z* is relevant; the fact that S represents$_D$ *z* is not relevant in this way. Analogously, if I have a map of Toronto, this should help explain not only my ability to navigate from Union Station to the Eaton Centre, but from any point in the city to any other point. This is what I aimed to capture with BEHAVIOR and ROLE, which relativize the attribution of representations not to the performance of specific tasks, but to feature-related behavior more generally. Given Harding's interests in operationalization for particular NLP applications, incorporating task-relativity is not a serious fault, but such relativity is best avoided in a general account of representation such as the one I am proposing.

A final difference is that Harding defends a misrepresentation criterion, whereas I do not. Like Shea (2018), her motivation is that representation should explain behavioral success, while misrepresentation should explain failure. As I explained in section 2.3, this criterion is too strong, since the same representation embedded in different algorithms can equally enable either robust success *or* robust failure. Relatedly, a representation which is *mostly* inaccurate can enable robust success within a particular task domain. Consider a map of Toronto which correctly depicts the highways but incorrectly depicts everything else. A driver traversing Toronto using only the highways will be just as successful as one with a fully accurate map. Harding might invoke task specificity to dodge this concern, but I have already provided some reasons to avoid such invocations. The result is that representation and misrepresentation *alone* do not explain anything, and so we need not include misrepresentation in a list of criteria designed to be diagnostic of representational status in the first place.[18]

## 3. LLMs as look-up tables

I am arguing that LLM behavior is often the result of representation-driven information processing. As noted, this claim would be trivial if representations were mere causal intermediaries between model inputs and outputs. My goal in the previous section was to provide a characterization of representation which is substantial enough to avoid the charge of triviality. Another way my argument would be rendered uninteresting is if there were no viable alternative explanations of LLM behavior. The goal of this section is to describe one powerful alternative, and then to show why it fails. This renders the representational strategy the best one available, since there are no other serious contenders in the offing. Section 4 explores in some detail how that strategy can be pursued. But first, the alternative.

---

[18] Another worry is that *knowledge* is a representational mental state which, as a conceptual matter, cannot misrepresent (since knowledge is factive). But presumably we would not want to conclude that knowledge is non-representational for that reason.

### *3.1. Look-up tables*

Large language models are neural networks. Neural networks are functions from inputs to outputs. Functions can be realized by various kinds of algorithm. Some algorithms, including many of those implicated in human cognition, involve the processing of information-bearing representations. But any finite function realizable by representation-based processing can also be realized by a "look-up table", in which the input-output pairs which characterize the function are explicitly encoded into the parameters of the processor.[19] Intuitively, a look-up table is a list of inputs matched with appropriate outputs. Given an input, the output is retrieved, rather than generated.

Neural networks are sometimes viewed as look-up tables, with input-output pairs encoded into the learned parameters of the network. On this view, "[e]xperience puts neural machines in enduringly different states (rewires them to implement a new look-up table), which is why they respond differently to inputs after they have had state-changing (rewiring) experiences)" (Gallistel & King 2009: 94-95). LLMs are neural networks with billions or trillions of parameters, and so perhaps they can be understood as encoding massive, tremendously complex look-up tables. The enormous size of such tables should not be underestimated, especially since it has been shown that neural networks can encode more features than they have dimensions via "superposition" (Elhage et al. 2022).[20] With sufficiently large and diverse tables, LLMs could plausibly produce articulate text and perform well on some cognitive tests solely in virtue of having encoded (or as is often said, "memorized") appropriate patterns from the training data.[21] If so, there would be little reason to suppose they have also developed representations (in my sense) of features of that data, since these would serve no functional role over and above that played by the look-up table.

### *3.2. LLMs as look-up tables*

---

[19] Look-up tables have played a significant role in the philosophy of cognitive science, most notably in Searle's "Chinese Room" thought experiment, where a monolingual English-speaking homunculus implements a Chinese conversation-function using a look-up table (Searle 1980), and in Block's hypothetical scenario involving a look-up table designed to mimic the linguistic behavior of his aunt Bertha (such a system is now called a "Blockhead") (Block 1981). These thought experiments were meant to shed light on the question of whether an appropriately programmed AI system might understand language or be intelligent, but I am interested in the more fundamental question of whether LLMs act according to look-up tables in the first place. See Grzankowski (2024) and Millière & Buckner (2024a) for discussion of look-up tables in relation to LLMs.

[20] Superposition is a type of distributed representation (Hinton et al. 1986; Templeton et al. 2024). Although superposition was originally thought to apply to "features" in the sense of generalized representations of the training data, Henighan et al. (2023) show that large models can also encode *data points* (input-outputs) in superposition. They claim that model overfitting is characterized by data-point superposition, while model generalization is characterized by feature-superposition.

[21] Zhang et al. (2021) show that the effective capacity of deep convolutional neural networks is often sufficient for memorizing a massive set of completely random data. See also Hartmann et al. (2023).

A number of recent papers characterize LLMs along roughly these lines. First, Bender & Koller (2020) claim that in order to respond coherently to evolving and unfamiliar inputs, an LLM "would need to memorize infinitely many stimulus-response pairs" (5193). In other words, it would require an infinitely long (and appropriately diverse) look-up table. Second, Bender et al. (2021) describe an LLM as "a system for haphazardly stitching together sequences of linguistic forms it has observed in its vast training data, according to probabilistic information about how they combine" (617).[22] This claim admits of two different interpretations. It could just be the trivial claim that LLMs generate outputs by combining linguistic tokens according to their distributional properties in a text corpus. That is just a description of the task being solved by the models, and leaves open the question of which sort of algorithm is employed to solve it. However, it could also be the claim that LLMs generate outputs by drawing stochastically ("haphazardly") from a look-up table learned through observations of token combinations in the training set. This is not a claim about the task, but about the algorithm. Accordingly, it is this interpretation which poses a challenge to my conclusion in this paper.[23] Third, Dentella et al. (2024) contend that LLMs produce outputs by "predict[ing] some of the fossilized patterns found in training texts" (7) and that they "do not go beyond reproducing input patterns" (7). In other words, LLMs memorize patterns of input from training data, and parrot those patterns back to the user during inference. This is just another way of describing a machine-learned look-up table.[24]

Systems based on look-up tables are characteristically *unproductive* – they "only give back what has been already put in" (Gallistel & King 2009: 93). In other words, they cannot generalize to inputs unlike those they have previously observed. Analogously, a multiplication table which provides answers to every multiplication problem from 1*1 to 12*12 cannot tell you the answer to 13*12. However, a system with compositional representations of numbers and algorithms for combining them should fare better. To test whether LLMs are acting on the basis of memorized associations or generalizable representations, we can present them with out-of-distribution inputs and see how they respond. Given their aforementioned size and breadth of learning, it can be difficult to know whether they have encountered a particular test before. However, a number of compelling recent papers have designed tests with this worry in mind, and their results strongly suggest that LLMs do *not* always rely on look-up tables when generating responses. I will argue in section 4 that such tests are not necessarily *diagnostic* of representation-based processing, as they are often taken to be, but that they are sufficiently suggestive as to render the look-up table view highly implausible in the cases discussed.

### *3.3. Case studies*

---

[22] This claim about the sort of processing in which LLMs engage is part of a larger argument that LLMs are unable to understand language or learn meaning. Discussing that larger argument is beyond the scope of this paper.

[23] Bender et al. also claim that LMs lack a "model of the world". Since this is a claim explicitly about the *content* of their representations, I do not directly consider it here, although some of the empirical evidence adduced in the next subsection bear strongly against it.

[24] See Shanahan (2023) and Titus (2024) for further arguments in favour of something like the look-up table interpretation.

### 3.3.1. Othello-GPT

In the first, widely-discussed study, Li et al. (2023) trained an autoregressive GPT on millions of transcripts of the board game Othello. The transcripts consisted entirely of move sequences (eg. A4, E6, B3, etc.) with no information about the rules of the game or the properties of the board. In the control condition, Othello-GPT was provided with a sequence of moves and prompted to continue it, with success defined by the proportion of legal moves taken. The model proved highly effective, achieving a success rate of 99.99% at its best. To check whether the model had merely memorized move sequences from this set, the authors retrained it on a custom "skewed" dataset which was missing a quarter of the game tree, meaning that the model had never seen a large number of possible board configurations. They then tested the model on a standard, non-skewed dataset, and it proved just as successful as the control (99.98% legal moves).

If Othello-GPT were relying on a look-up table, it would be mysterious how it managed to perform so well in the experimental condition, since in that case it had no access to the relevant "stimulus-response pairs". It could not merely be stitching together sequences of forms it observed in its training data, since many of the sequences on which it was tested were not present in that data. Accordingly, the look-up table interpretation of Othello-GPT's behavior is implausible.

### 3.3.2. Model of colour space

In the second study, Patel & Pavlick (2022) tested a pre-trained GPT on its ability to make inferences about the structural properties of two domains, namely the domains of colour and space. For brevity I will focus on colour. The authors employed an in-context learning paradigm where the model was provided with sixty training examples and then tested on unseen cases. In the control condition, training examples consisted of pairings of RGB codes with their corresponding colour names (eg. RGB: *[255, 165, 0]*, Answer: *orange*) drawn from a limited part of the colour spectrum. At test time, models were prompted with RGB codes from a different part of the spectrum and asked to produce the corresponding colour names. The model achieved 34% Top-3 accuracy on this task (where the correct answer was within the three most probable answers provided by the model) – significantly above chance (13%),[25] and possibly artificially low given that a mark of "correct" required an exact string match (so that, for instance, a model which output "wine" when the correct answer was "dark red" would be marked as incorrect).

To ensure that the model had not merely memorized colour-code pairings from its training set, they tested the model in a "rotated" condition where the colour-code pairs were systematically permuted so as to be isomorphic to the original pairs. In other words, all individual colour-code pairs were different in an absolute sense, and so unlikely to be present in the training data, but the structural relations between them (i.e., their relative distance and directional properties in colour space) remained the same. The model performed equally well in the rotated condition as in the control (36% Top-3 accuracy). If this model were acting via a look-up table, it would be very difficult to explain its generalization ability in the novel, rotated context. The natural alternative explanation is that it had learned a representation of the structure of the colour space which enabled it to infer the "location" of new colours (really, colour names) based on information about others.

---

[25] They don't explicitly define a chance value, but 13% is the model's success rate in a condition where the colour-code pairs were permuted randomly, so it should serve the same purpose.

These are two studies among many which strongly suggest that LLMs are not just stochastic look-up tables.[26]

### *3.4. Objections*

There are a number of ways one might push back against my claims in this section.

#### 3.4.1. Fuzzy hash tables and approximate retrieval

Look-up tables with explicitly encoded input-output pairs are unproductive, and so cannot adequately respond to unfamiliar inputs. However, some have suggested that due to their probabilistic nature, LLMs they are less like multiplication tables and more like fuzzy hash tables (Saba 2024), where inputs with shared features are bucketed together and associated with similar outputs. Whereas standard look-up tables facilitate exact retrieval, which is incompatible with out-of-distribution generalization, fuzzy hash tables facilitate *approximate retrieval* (Kambhampati 2024), which is not.

To illustrate, compare exact and approximate machine-learned translation systems. An exact English-French translator deterministically pairs inputs with outputs: cat ↦ *chat*; kitten ↦ *chaton*; and so on. Such a system would be unable to translate the unseen word, "kitty". By contrast, an approximate English-French translator pairs inputs with outputs based on shared input features. Suppose such a system defines buckets using sets of bigrams, as follows: {ca, at} ↦ *chat*, {ki, it, tt, te, en} ↦ *chaton*¸ etc.[27] Given the new word, "kitty", the system buckets it based on bigram similarity. In this case, its constituent bigrams {ki, it, tt, ty} match best with the second bucket, and so the word would also be translated as *chaton*, demonstrating generalization to out-of-distribution inputs.

Even if this provides an accurate depiction of how LLMs produce outputs, it does not provide a genuine alternative to the representational strategy, since the buckets into which novel inputs are placed are plausibly considered learned representations of input features. The system is comparing inputs to stored representations (in this case, of bigram features) and computing outputs based on their match with those representations. Analogously, a system trained to make grammaticality judgements might learn to bucket inputs based on their similarity to stored representations of syntactic patterns. This is not an alternative to representational explanation, it is just a way of fleshing-out how it might work.

#### 3.4.2. LLMs as mere compression algorithms

---

[26] For some others, see Garg et al. (2022) and Raventós et al. (2023) for evidence that LLMs can learn out-of-distribution function classes from in-context examples, Lampinen et al. (2023) for evidence that LLMs can make generalizable inferences to unseen links in causal graphs, and Misra & Mahowald (2024) for evidence that LLMs learn representations of abstract grammatical rules which enable generalization to rare syntactic structures.

[27] Although note that buckets are typically defined as regions of feature space rather than exact feature lists.

A related objection is suggested by a metaphor for LLMs proposed by Chiang (2023), according to which LLMs are akin to lossy compression algorithms, or "blurry JPEG[s] of all the text on the Web". It is in virtue of this compression that LLMs don't simply memorize exact patterns of text from their training corpora: "you can't look for information [in an LLM] by searching for an exact quote; you'll never get an exact match, because the words aren't what's being stored". For Chiang, compression explains the appearance of novelty and generalization in the outputs of LLMs, but does not underly any actual capacity to reason or understand the tasks they perform.

Chiang is certainly right to compare LLMs to compressors, since it is well known that an optimal sequence predictor for some data is an optimal compressor for that data (Delétang et al. 2024; Hutter 2005). A system which maximizes the log-likelihood of its data (i.e., which minimizes cross-entropy loss, as LLMs do) is *ipso facto* a system which achieves the minimum description length of its predictive model: "the best way to make predictions is based on the shortest (≙ simplest) description of the data sequence seen so far" (Hutter 2005, 30). Regardless of the implications of Chiang's point for the prospect that LLMs can reason or understand, it poses no challenge to the idea that they can form and use representations. Indeed, as Trott (2024) observes, "this property of 'compressing input into lossy but useful representations with which to make future predictions' looks a lot like certain theories of cognition", namely representational theories. In other words, the compression view isn't an alternative to the representational one. Rather, compression is a candidate *mechanism* by which LLMs form the representations which guide their future behavior.

In this section, I have explained what it would mean for LLMs to be look-up tables, described how this view predicts LLMs will behave, introduced two empirical results which disconfirm those predictions, and responded to some objections. Note that my conclusion is not that LLM behavior is never driven by deterministic table look-up: sometimes it certainly is (Zhang et al. 2021), and indeed look-up tables (and their psychological counterpart, rote memorization) are central to the behavior of complex systems of all sorts, from digital computers to human brains.[28] The claim, rather, is that LLMs do not *always* act according to look-up tables, and that in such cases representation-based information processing is likely to be at work. My goal, however, is not merely to uphold the in-principle applicability of a representational approach, but to describe and defend the viability of some practical techniques for pursuing it. This is the topic of the next section.

## 4. In search of representations

I began this paper by providing a characterization of representations which capture their special explanatory role in cognitive science. This characterization includes four jointly sufficient conditions: INFORMATION, EXPLOITABILITY, BEHAVIOR, and ROLE. After elaborating on the conditions and explaining how my characterization differs from that proposed by Harding (2023), I considered an alternative, non-representational strategy for explaining LLM behavior in terms of table look-up. This strategy predicts that LLMs will fail to generalize to truly out-of-distribution tasks, and I described two studies – Li et al.'s (2023) Othello-GPT study, and Patel

---

[28] Kiyomaru et al. (2024) show that memorization is more common in larger LLMs and with text that appears frequently in the training data, and Menta et al. (2025) demonstrate that attention blocks in later layers are typically responsible for memorization, whereas earlier layers are key for generalization.

& Pavlick's (2022) study on LLM representations of colour space – which demonstrate robust generalization abilities. Lastly, I considered some objections which attempted to modify the look-up table view such as to account for these abilities, and argued that both of them are ultimately just ways of fleshing out the representational strategy rather than genuine alternatives to it. In this section, I argue that the presence or absence of representations is inescapably underdetermined by behavior alone, necessitating the use of techniques for empirically investigating LLM internal states, techniques which I go on to describe and defend.

### *4.1. Diagnostic behavior?*

The discussion so far has suggested that there is something special about model behavior which is flexible and generalizable – in other words, robust – such as that exhibited by Othello-GPT. One might expect that this sort of behavior is *diagnostic* of the presence of representations (eg. as claimed by Dasgupta et al. [2019]), since it is often difficult to imagine how a system governed only by stimulus-response pairings or some similarly inflexible substructure could take such generalizable actions. Note that this is not the same as accepting BEHAVIOR as a *condition* on representation. The perspective I am concerned with in this subsection is one which takes flexible behavior alone as *diagnostic* of the presence of representations.

#### 4.1.1. Maha's map

To illustrate the perspective with an analogy, imagine Maha is driving home from work along a memorized route when her partner calls and asks her to pick up the kids from school on the way. Suppose that Maha has never picked up the kids from her current location. To succeed at this task, Maha would appear to need some sort of map (a spatial representation) of the neighbourhood, whether on her phone or in her head, in order to change course and reach her new destination.[29] If she could only follow routes which she remembers having previously taken, she would be stuck. However, I argue that the presence or absence of representations is underdetermined by Maha's mere behavior, and analogously, by the behavior of LLMs; there is no such thing as a *diagnostic behavior*.

Suppose Maha arrives home with the kids, and this is all we know about her trip. She could have used a map in the manner just described. However, she could also have tried every possible route from her original location until she reached the school, thereby solving the navigation problem by brute-force. Or, she could have followed some sort of ecologically valid heuristic, such as "take every right turn", which would lead her to the school from any starting point. Acting according to a rule like "try every possible path" or "take every right turn" is not acting via a representation, at least not in the sense of section 2.

This point is worth clarifying.[30] As in action theory we distinguish actions done *according to* reasons from those done *for* reasons (Piñeros Glasscock & Tenenbaum 2023), we must distinguish actions done *according to* (or *in conformity with*) rules from those done *in virtue of* rules, where representing a rule is a necessary condition on the latter. One might habitually act in a way that conforms with a particular rule, without acting in virtue of it; for instance, you might jiggle your leg when anxious, without aiming to follow the rule, "jiggle one's leg when anxious". Analogously, Maha might follow the right-turn rule out of habit without acting in virtue of it – perhaps she finds right turns more comfortable and so defaults to making them when she's lost. In such a case, she does not represent the right-turn rule (she doesn't believe it will lead her to the school), but she nevertheless conforms to it. The point is simply that from her behavior alone we can't tell the difference.

---

[29] See Tamir & Shech (2023) for discussion of a similar example. They emphasize that the key feature of map-like representations which enables flexible behavior is how the information they contain is *structured*. Behrens et al. (2018) make a comparable point.

[30] Thanks to an anonymous review for encouraging me to do so.

Moreover, supposing Maha *fails* to arrive home with the kids, this does not entail that she *lacks* a relevant representation either. For instance, she could have a map, but an inaccurate one; even slight inaccuracies can have significant implications for downstream performance. Or, despite having a perfectly accurate map, she might still fail due to environmental conditions, for instance because construction has blocked all possible paths or because she is distracted. As emphasized above, behavioral success is not always explained by the use of an accurate representation, nor is failure always explained by its absence.

#### 4.1.2. Performance, competence, and heuristics

In these scenarios, Maha's navigational performance is underdetermined by her possession of a map (she can do well or poorly with or without one), and vice-versa. This is an instance of the more general principle of performance-competence underdetermination (Chomsky 1965), where performance is outward behavior, and competence is "a system's underlying knowledge: the internal rules and states that ultimately explain a given capacity" (Firestone 2020: 26564), and where these rules and states are (or are often) representational.[31] As Harding & Sharadin (2024) observe, "[t]his distinction is frequently elided in evaluations of ML model capabilities" (11):[32] good performance (for instance, on benchmarking tasks [Srivastava et al. 2023]) is often taken to be sufficient for competence,[33] and bad performance for a lack thereof. However, a system may perform well at a task despite lacking deeper competence (for instance, by relying on heuristics or other tricks, such as Maha's "right turn" rule), or it may perform poorly despite having such competence (for instance, because it is constrained in some extrinsic way, eg. by construction in Maha's case).

One complication is whether heuristics themselves can play a role in representational explanations. For example, McCoy et al. (2019) showed that the language model BERT relies on several heuristics when performing natural language inference (NLI), including the "lexical overlap heuristic", according to which a premise entails all hypotheses constructed from words in the premise.[34] Presuming that the model had internal states which carried exploitable information about lexical overlap in the training set and that this information played a mechanistic role in the model's robust NLI performance, did the model not have a representation with content like: "premises entail all hypotheses constructed from words in the premise"?

---

[31] See Millière & Rathkopf (2025) and Grzankowski et al. (2025) for application of the performance-competence distinction to LLMs.
[32] Firestone (2020) and Pavlick (2023) make similar points.
[33] Such that, eg., success on reasoning tasks demonstrates an ability to reason.
[34] Where hypotheses are statements that the model must judge as being either entailed by or contrary to a previously provided statement, the premise.

The ambiguity here lies in the notion of "robustness". Heuristics may be valid in a narrow or a wide set of environments, where narrow validity entails low robustness and wide validity entails high robustness. If Maha were following the right-turn rule, but one of the relevant turns was blocked, or if she were in a different city altogether, her performance would seriously degrade and thereby demonstrate a lack of robustness to different initial conditions and perturbations. Likewise, if BERT were following the lexical overlap heuristic when tested on a dataset in which the heuristic is invalid, it would fail the inference task (as also demonstrated by McCoy et al.). In this sense, heuristics do not explain modally robust task performance, and so invoking them fails to satisfy BEHAVIOR.

But this is too quick: there are also many environments in which heuristic-guided behavior succeeds, for instance when Maha begins her route from different starting points in the same city with no construction, or when BERT is tested on other common NLI datasets in which the lexical overlap heuristic remains valid. Evaluations of robustness are thus task- and environment-specific, and so whether BEHAVIOR is satisfied in a particular case will partially depend on the task and environment in question. Robustness is not a well-defined notion to which we can plausibly set hard, universally applicable limits. Representational explanations of LLM behavior are thus somewhat context-dependent, but so are representational explanations of *any* system's behavior. Indeed, context-dependence (often couched in *ceteris paribus* clauses) is a normal feature of all explanations.[35,36] If, for example, we want to explain how Saul is so good at making friends, testing his friend-making ability on jellyfish, or when he is asleep, is a non-sequitur. We want to know how he makes *human* friends in *typical* scenarios; that is the domain in which his behavior is robust, and the one which calls out for explanation. Likewise, in the NLI case, what we are trying to explain is how BERT manages to perform so well on the relevant datasets. If instead we are using NLI tasks to test BERT's *understanding* of language more generally, the fact that it fails on some datasets becomes much more significant.[37] But this is precisely what we should expect: different questions demand different answers.

### *4.2. Mechanistic interpretability*

To move beyond behavior, we need tools and techniques which allow us to systematically investigate not only *what* LLMs do in response to input, but *how* those responses themselves are generated. Researchers in the new field of "mechanistic interpretability" ("MI"; Olah et al. 2018, 2020; Millière and Buckner 2026) have developed a number of ingenious techniques for probing and intervening on the activations and weights of deep neural networks (DNNs). These techniques open the "black box", allowing us to directly test hypotheses about how the behavior of these networks arises from their internal processing. Most relevantly for our purposes, MI techniques provide natural ways of testing whether INFORMATION,

---

[35] Cf. Boge (2024).

[36] Except perhaps for explanations in terms of fundamental laws which are meant to hold universally (though see Cartwright [1999] for doubts about that).

[37] See Tamir & Shech (2023) for discussion of reliability and robustness in measures of machine understanding.

EXPLOITABILITY, BEHAVIOR, and ROLE are satisfied in particular cases, and so whether an LLM is relying on representations in those cases.[38]

#### 4.2.1. Interpreting INFORMATION and EXPLOITABILITY

INFORMATION requires that R carry information about *z*, and EXPLOITABILITY requires that the information R carries about *z* is exploitable by S, where (in our case) S is the LLM under investigation. The set of MI techniques known as "probing" offer one simple way to test whether an R carries the relevant information, and whether it is exploitable, by testing whether that information can be decoded from it (Alain & Bengio 2018).[39] Probes are typically small, linear, multi-layer perceptrons (MLPs) trained using supervised learning. They take as input some state R of the target network S – eg. a set of hidden-layer activations – and output predictions which are compared to ground truth labels, with the resulting error signal used to update their parameters, before the process repeats.[40] Ground truth is determined by whether the feature of interest *z* was present in the input to the target network.

For example, suppose we want to know whether the first hidden layer of an English-French translation model carries information about the language of the network's input (i.e., about whether it is in English or French). Since we can't merely look at the layer's activation values and intuit the answer, we instead use those values as input to a probe. The probe makes a prediction (at first, a random guess) about whether the original input was in English or French, and its parameters are updated in light of the resulting error. The process is iterated with new activation values from the same layer until the probe's prediction error stabilizes (Fig. 1). If it stabilizes at a sufficiently low value, we may conclude that information about the language of the input (*z*) is present in the activation values (R) of the first hidden layer of the target network (S).[41] In other words, some properties of the activation values are correlated with language-relevant features of the input. If that correlational information were absent, the probe's error would remain high.

#### 4.2.2. Complications

---

[38] Harding (2023) provides detailed and rigorous operationalizations of conditions similar to INFORMATION and EXPLOITABILITY specifically in terms of MI methods.

[39] Probing is a good illustrative example since it has been around for some time and is reasonably well understood, though for some purposes it has been surpassed by other methods like dictionary learning which are easier to apply at large scales (Templeton et al. 2024). That said, Millière & Buckner (2024b) note that dictionary learning "lacks robust theoretical grounding" compared to other interpretability methods, and there is evidence that probing is a more reliable and generalizable means of detecting representations in LLMs (Wu et al. 2025), which further supports the choice of probing as an illustration of MI methodologies.

[40] Many factors will guide our selection of R: the particular model being studied, our previous knowledge of how it works, our research question, etc. R could be activations from a single layer or across multiple layers, it could be a series of weight values, it could be a specific model component such as an attention head, etc.

[41] There are different methods for deciding what constitutes a "sufficiently low value", and for quantifying the *amount* of information R carries. See Harding (2023) for discussion.

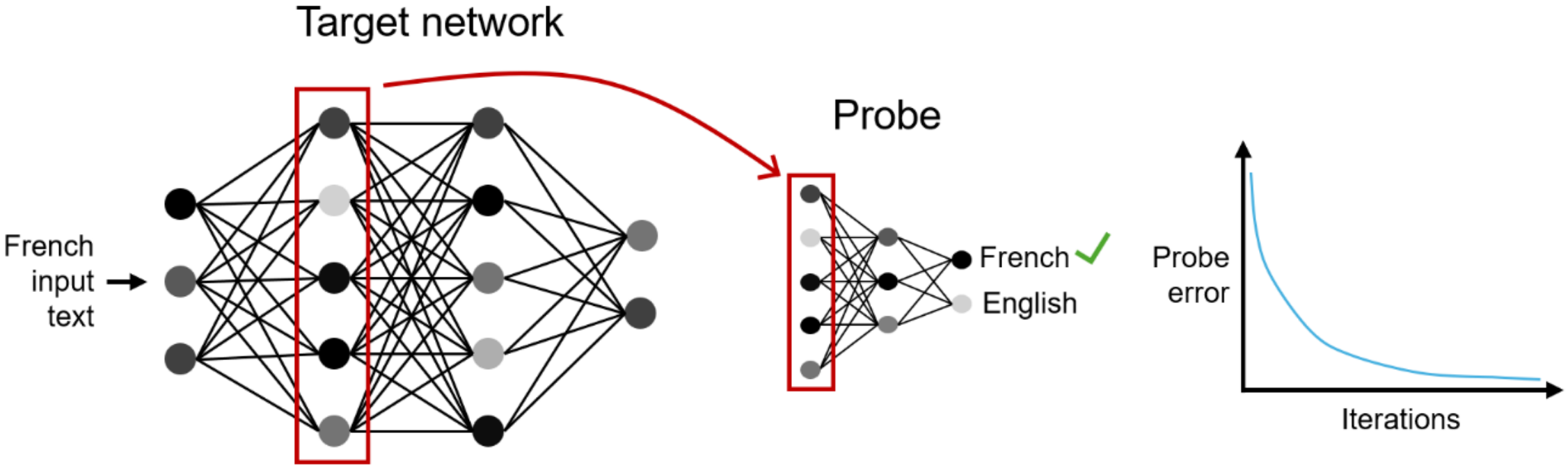

**Fig. 1** Toy illustration of probing technique. The target network receives an input of vectorized text and processes it. The activation values at the first hidden layer are retrieved and fed as input to a probe, which is less complex than the target (indicated visually by their relative sizes). The probe outputs a prediction about whether the target network's input was in English or French, based solely on the hidden layer activations. In this example, the probe predicts "French" with some degree of confidence less than 1, and the resulting error is backpropagated. Over many iterations of this process, the probe's prediction error is reduced significantly, indicating that information about the language of the input text is present in the activations of the first hidden layer of the target network

This is an idealized picture of how probing works. There are a number of complications which must be dealt with in actual cases, but they do not seriously compromise the value of the technique. First complication: the mere fact that some information is decodable by a probe does not entail that it is exploitable by the downstream layers of the target network; INFORMATION might be satisfied while EXPLOITABILITY is not. If the probe is more complex than the target,[42] for instance, it may be able to learn a function which is unlearnable for the target. Analogously, if I get stuck while following a series of clues on a scavenger hunt, the mere fact that *someone* can successfully follow the clues does not meant that *I* can, since they could simply be cleverer than me. This is why probes are typically small, linear MLP networks, since any function learnable by such networks is also learnable by a highly complex non-linear deep neural network.[43,44] Returning to the analogy, if a *child* can successfully follow the clues, surely I can as well.

This suggests a second complication which is the inverse of the first: if the child *cannot* follow the clues, that does not mean that *I* can't. In other words, while the success of a probe is good evidence for the presence of some particular information, the failure of a probe is poor evidence for the absence of that information. Hence, we ought to employ many different probes – Harding (2023) suggests that we select as probes the smallest, most successful models within each model family, where "success" is inversely proportional to the model's error rate on the probing task, and "model families" are set by, for instance, the size and number of hidden layers, the rank of linear transformations, etc.[45] If at least one carefully selected probe is successful, we can conclude that the information is decodable, and thus exploitable, by the target network. In such a case, INFORMATION and EXPLOITABILITY are satisfied.

[42] If, eg., it has more parameters, more layers, more non-linearities, etc.
[43] See also Harding's (2023) discussion of the "size" and "linearity" constraints on probe selection.
[44] Also, it is likely that linear representations are common even in highly complex networks with many non-linearities: "[n]eural networks are built from linear functions interspersed with non-linearities. In some sense, the linear functions are the vast majority of the computation (for example, as measured in FLOPs). Linear representations are the natural format for neural networks to represent information in!" (Elhage et al. 2022).

Levinstein & Herrmann (2024) propose a third difficulty with probing: we can seldom know whether a probe is identifying properties of the target network which carry information about the feature of interest *z*, or about some other feature *z** which spuriously correlates with it in the original training set. In their case, they trained the language model LLaMA 1 (30b) on six labelled datasets of true and false statements and found that probes were effective at distinguishing the truths from the falsehoods given a set of LLaMA's intermediate activations.[46] This suggested that the probes found "representations of truth" (§5.2) within those activations. However, when tested on negated versions of the original statements, the same probes' classification accuracy plummeted, some of them to below chance. From this, the authors conclude that:

> Since the probes failed to do well on [the negated datasets] even after training on all positive analogs along with other negative examples, it's likely the original probes are not finding representations of truth within the language model embeddings. Instead, it seems they're learning some other feature that correlates well with truth on the training sets but that does not correlate with truth in even mildly more general contexts. (§5.4)

The upshot is that interpreting the results of probing experiments is less straightforward than it initially appears.[47]

While Levinstein and Herrmann raise a legitimate concern, it is no different in principle from the concern about heuristics discussed in 4.1.2. In both cases, the network – whether target or probe – learns a task by picking up on features different from those we intuitively expected: not entailment, but lexical overlap; not truth, but something correlated with it; not *z* but *z**. Nonetheless, probes remain useful for at least two reasons.

First, if all attempted probes are unsuccessful, we gain strong (though defeasible) evidence that neither information about *z* nor about *z** is present, and so that INFORMATION is not satisfied. In such cases, the target network is likely relying on a non-representational strategy such as a look-up table.

Second, if a probe is successful, we still learn that R carries correlational information about *z*. It may just be that R is most directly correlated with *z** rather than with *z* itself, and so reduces uncertainty about *z* indirectly. In the same way, neural firing in V1 directly indicates the presence of particular patterns of retinal stimulation, yet the information it carries is processed as information about the presence of objects in the environment. Not all correlations are spurious.

---

[45] If we are still worried about overpowered probes, methods such as Hewitt and Liang's (2019) "control tasks" can be used to filter out those which are capable of achieving low error through brute force memorization.

[46] The true-false statements were from Azaria & Mitchell (2023). The probes' binary classification accuracy ranged between 72.2% and 97.3%.

[47] Cf. Belinkov (2022).

#### 4.2.3. Interpreting BEHAVIOR and ROLE

Whether an R counts as a "representation of truth" or of anything else depends not only on its exploitable information, but on whether the wider system S exploits that information to enable robust BEHAVIOR, and whether R plays a mechanistic ROLE in that behavior. Helpfully, MI techniques also provide natural ways of testing whether BEHAVIOR and ROLE are satisfied.

R plays a mechanistic role in S's robust *z*-related behavior if it causally contributes to the mechanisms driving the behavior (Craver 2007). Assuming an interventionist account of causation, we can test for R's causal contribution by intervening on it – either by changing or eliminating it – and observing the resulting behavioral differences (Woodward 2003). For BEHAVIOR and ROLE to be satisfied, the intervention should modify R so as to corrupt or otherwise alter the information it carries about *z*, with the result that S's *z*-related behavior degrades or changes in some way. For instance, if we are investigating how an LLM forms grammatical sentences, and we know R encodes part-of-speech information, ablating R should lead to greater difficulty in identifying parts of speech. If instead it only leads to difficulty in, say, distinguishing dogs from horses, or to no difficulty at all, R does not play a mechanistic role in S's robust *z*-related behavior, and so it does not play the representation-role in the relevant context (although it may do so in dog/horse-related contexts).[48]

A number of methods have recently been developed for intervening on the internal states of deep neural networks such as LLMs.[49] For concreteness, I will briefly describe one technique before defending the general strategy from some objections. The simplest technique plays on the increasingly popular idea that DNNs encode features as linear directions in activation space (Elhage et al. 2022; Park et al. 2023). Intuitively, a set of activations R carrying information about feature *z* will "point" in the "*z*-direction" (Fig. 2a). If one knows the direction of another feature – call it *k* – one can add a linear vector $\vec{v}$ to the original activation so that it "moves" from the *z*-direction to the *k*-direction (Fig. 2b). In other words, R carrying information about *z* becomes R* carrying information about *k*. Note that, despite the abstract terminology of "directions" in "activation space", this intervention corresponds to a causal change in R, since activation values correspond to states of the physical processor. Ultimately, it is those states which play a mechanistic role, and those states are precisely what get changed by an intervention like this. If, as a result of this intervention, the network S's *z*-related behavior degrades or otherwise changes, we can feel confident that BEHAVIOR and ROLE are satisfied.[50]

---

[48] If this seems implausible, consider the fact that a paradigm representation – a realistic sculpture – can serve as a representation in one context (say, when viewing it in an art gallery), while playing a non-representational role in another context (say, when using it as a paperweight).

[49] Including activation patching (Heimersheim & Nanda 2024), iterative nullspace projection (Ravfogel et al., 2020, 2024), activation steering (Radford et al. 2016) and knockout/ablation (Olsson et al. 2022). Some of these techniques also get discussed under the headings of "concept erasure" (Belrose et al. 2023) and "representation engineering" (Zou et al. 2023).

[50] Though see section 4.2.4 for potential complications with this picture.

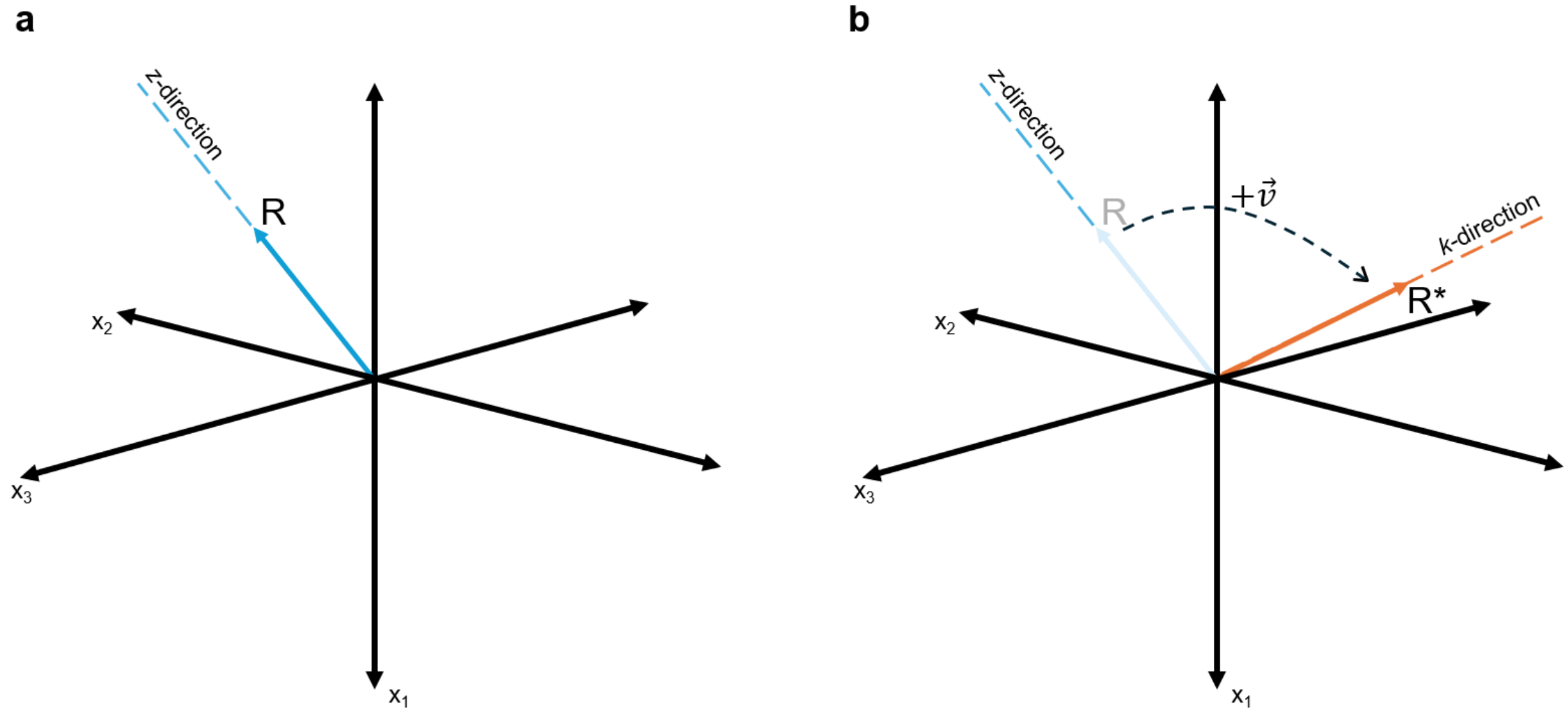

**Fig. 2** Toy illustration of basic intervention technique in a 3D activation space. **(a)** A set of activations R carrying information about feature *z* points in the "z-direction" of activation space. Note that perfect alignment between R and the *z*-direction is an idealization. **(b)** Adding a linear vector $\vec{v}$ to R changes its values to those of R*, which points in the *k*-direction. This vector addition operation is possible since R, a set of activations, is just a vector of activation values. Note that in reality, the activation spaces of large language models contain hundreds of thousands of dimensions, but the general process remains the same

Further research on Othello-GPT provides a concrete example of this technique at work. Using the model weights provided by Li et al. (2023), Nanda et al. (2023) trained a linear probe to classify the state of each board tile at a timestep, where tiles are classified as either "mine", "yours", or "empty".[51] The probe's success indicates that the board state is linearly decodable from the target network, i.e., that the network S has internal states R which carry exploitable information about the board state *z*.[52] Hence, INFORMATION and EXPLOITABILITY are satisfied. To evaluate the causal role of these states, the authors add a linear vector to the residual stream of each layer of the target network which "flips" a particular tile from "yours" to "mine" or vice-versa, and then compare the predictions of the model for the next likeliest move both before and after the intervention.

They found that the model's predictions were consistent with the alteration in its purported representation of the board: flipping a tile from "yours" to "mine" rendered the predicted moves consistent with the "mine" board state. This demonstrates that the information-carrying states of the target model play a causal role in the mechanisms underlying the model's Othello-related behavior, as evidenced by the changes in its predictions. These states thus play the representation-role, and can feature in representation-based explanations of Othello-GPT's behavior. For instance, Nanda et al. (2023) develop an account of how these representations figure to different degrees in the circuits underlying model behavior, finding that as individual Othello matches progress, the model often computes its next move *before* computing the board state; this is most likely because by the late-game, almost all empty tiles are legal moves, such that it is possible to predict legal moves without relying on a representation of the board. This explanation of the dynamics of Othello-GPT's behavior would be impossible without adopting a representation-based approach.

---

[51] This is a slightly different probing methodology than that pursued by the original authors, who trained the probe to classify amongst "black", "white" and "empty" tiles, and found that only non-linear probes were successful (Li et al. 2023).

[52] The same conclusion about linear decodeability in Othello-GPT is reached by Hazineh et al. (2023).

Templeton et al. (2024) show how MI techniques can be scaled up to identify and intervene on representations of many different features. They use sparse autoencoders (SAEs) – neural networks under strict regularization constraints designed to encourage the formation of distinct, interpretable features – to learn a "dictionary" of basic features which causally contribute to the behavior of the Claude 3 Sonnet language model. They applied SAEs to residual stream activations halfway through the model, and found patterns of activation corresponding to a number of interpretable features, including the Golden Gate Bridge, transit infrastructure, and brain sciences. To test for mechanistic role, they perform *feature steering*, which involves clamping the activation values for particular features to artificially high or low values at inference. For example, clamping the Golden Gate Bridge feature to 10x its original activation value causes Claude to mention the bridge when it previously would not, and even to self-identify *as* the bridge. Similar effects were observed for the other features: eg. whereas the base model outputs "physics" when asked what is the most interesting science, it outputs "neuroscience" when the brain sciences-feature is clamped to high values. Although detailed analysis is not possible here, these features appear to satisfy all four conditions on representation. More generally, dictionary learning and feature steering provide promising ways to develop large-scale representation-based explanations of LLM behavior going forward.

#### 4.2.4. Complications

Like probing, MI interventions may be difficult to interpret for at least two reasons. First, even if degraded or altered post-intervention behavior is diagnostic of mechanistic role, the *absence* of degraded behavior may not be diagnostic of a *lack* of mechanistic role, since the behavior in question could be causally overdetermined. An event is causally overdetermined if it has multiple independently sufficient causes (Bunzl 1979; Céspedes 2016). For example, if my house burns down due to a grease fire and a simultaneous lightning strike, the event of my house burning down is causally overdetermined, since either the grease fire or the lightning strike would have been sufficient for bringing it about. If one of these causes were absent, the outcome would be the same, even though both played a causal role. Analogously, the fact that an intervention on some activations produces no change in an LLM's behavior does not entail that they played no causal role in the mechanisms underlying the behavior, since it could have been overdetermined.[53]

---

[53] Indeed, causal overdetermination, or redundancy, is a ubiquitous feature of neural networks both biological (Mizusaki & O'Donnell 2021) and artificial (Morcos et al. 2018). Part of the point of regularization procedures such as dropout (Srivastava et al. 2014) and batch normalization (Ioffe & Szegedy 2015) is to encourage the development of internal redundancies as a way to avoid overfitting. See Mueller (2024) for related discussion.

Another possible complication, noted by Harding (2023), is that "*distinct representational contents* can have the *same representational vehicle*" (16, original emphasis). In other words, a given R might encode information about several different behavior-relevant features, and so if model performance degrades after intervening on R, it may be hard to pinpoint the particular representation – the particular vehicle-content pair – that is responsible.[54]

Some practical strategies are available for enabling effective interpretability despite these complications. In response to the second, Harding defines a "control condition" designed to allow us to determine whether contents overlap: "an intervention on the information carried by [an activation vector] about property $Z$ relative to the downstream system $S_{Dec}$ satisfies the control condition for property $Y \neq Z$ just in case it (approximately) preserves the predictions of probes for $Y$" (17). In other words, we can check whether an intervention has influenced multiple behavior-relevant representations (here, representations of $Z$ and $Y$) by assessing whether a probe trained to predict $Y$ can successfully do so after the intervention on the representation of $Z$. If it can, that suggests information about $Y$ is decodable separately from information about Z. If it can't, that suggests information about Z is inextricably tied to information about Y in virtue of having the same vehicle. This isn't ideal, but we still learn something important about how the system works using this method. More generally, it poses no problem for the representational approach writ large, only with a particular technique for implementing it.

A similar approach applies to the first problem. If we wanted to know whether the grease fire or the lightning strike caused my house to burn down, and we had a time machine, we could go back, prevent one of them from occurring, and see what happens. If the house burned in both cases, we would learn that the event was causally overdetermined. In the LLM case, we *do* effectively have a time machine, since we can simply run the model multiple times on the same inputs, each time testing for the causal role of a particular R. If the relevant behavior turns out to be overdetermined, that is itself an interesting result, and in no way interferes with our ability to provide a representation-based explanation. Moreover, causally overdetermined behavior is ubiquitous in the biological sphere as well – for example, a beaver's dam-building behavior may be simultaneously caused by an instinctive response to the sound of running water, as well as by an inclination to mimic its conspecifics. Learning that this is the case does not confound our ability to explain the beaver's behavior; rather it *constitutes* an explanation of the beaver's behavior.

Finally, the strategy of using MI interventions to detect representations can also be defended via a companions-in-guilt argument. *All* applications of interventionist theories of causation must grapple with overlapping and overdetermined causes – especially in dealing with complex systems such as artificial or biological neural networks – and with the concomitant task of developing control conditions like that described above.[55] Since interventionism is "more or less in line with how causation is understood in scientific practice" (Dijkstra & de Bruin 2016), the companions in guilt are *bountiful*. Hence, mechanistic interpretability is safe from in-principle difficulties, and is likely to enable a flourishing of representation-based explanations of LLM behavior in coming years.

---

[54] This is related to the notion of "polysemanticity", according to which "individual neurons in neural networks often encode information about multiple properties or concepts" (Millière & Buckner 2026).

[55] For instance, there is significant debate in cognitive neuroscience over how to interpret the results of transcranial magnetic stimulation (TMS) and other kinds of interventions, which follow a comparable methodology to the MI methods just described (see, eg., Robertson et al. 2003; Bergmann & Hartwigsen 2021).

## 5. Conclusion

Large language models are highly complex systems whose behavior on tests of cognitive ability often matches or exceeds that of humans. Sometimes this behavior is best explained by brute-force processes of memorization or stochastic table look-up. Other times, as I have argued, this behavior is best explained by the positing of subpersonal representations of precisely the sort posited to explain much human and animal behavior. In section 2, I offered a characterization of representation designed to capture their special explanatory role in cognitive science. In section 3, I argued that the key alternative explanatory approach – according to which LLMs are *merely* look-up tables – fails in some key cases. In section 4, I argued that behavioral testing alone cannot tell us with certainty whether a system is relying on a representation, but that we can become more sure by employing targeted intervention methods. The first method, probing, allows us to determine whether the LLM has internal states which satisfy INFORMATION and EXPLOITABILITY. The second method (really, a set of methods) allows us determine whether those states also satisfy BEHAVIOR and ROLE. I then described how all of these methods have been used to investigate Othello-GPT, such that we can feel very confident that some of its behavior depends on the processing of representations.

What I haven't done in this paper is provide a systematic account of *when* we should expect LLMs to rely on representations and when we should expect other sort of mechanisms be in play. I have also said nothing about how representational content is determined in LLMs,[56] nor about the broader metaphysical and epistemic status of their representations – whether they constitute concepts, beliefs, perceptions, knowledge, or understanding. However, there would be no point in investigating their status if we were not sure whether they had robust representations in the first place. This paper has argued that they do. The implications are yet to be worked out.

**Acknowledgements**

I would like to thank Jennifer Nagel, Sara Aronowitz, and Nate Charlow for extensive feedback and support, as well as audiences at the Dalhousie University Philosophy Department Colloquium, CogSci 2024, and the 2024 Descartes Lectures workshop.

No conflicts to declare.

---

[56] Though see Mollo and Millière (2026) for discussion.